\documentclass[letterpaper, 10 pt, conference]{ieeeconf}  

\IEEEoverridecommandlockouts                              

\usepackage{amsmath} 
\usepackage{color}
\usepackage{graphicx} 
\graphicspath{ {figs/} }  
\usepackage{float}
\usepackage{booktabs}
\usepackage{cite}
\usepackage{diagbox}
\usepackage{multirow}
\usepackage{algorithm}
\usepackage{algorithmic}
\usepackage{lineno,hyperref}

\title{\LARGE \bf                                                                                                                                                                                                                                 
HARP: Human-Assisted Regrouping with Permutation Invariant Critic for Multi-Agent Reinforcement Learning
}

\author{Huawen Hu$^{1}$, Enze Shi$^{1}$, Chenxi Yue$^{1}$, Shuocun Yang$^{1}$, \\ Zihao Wu$^{2}$, Yiwei Li$^{2}$, Tianyang Zhong$^{1}$, Tuo Zhang$^{1}$, Tianming Liu$^{2, *}$, Shu Zhang$^{1, *}$ \\ 
\thanks{$^{1}$Northwestern Polytechnical University, Xi'an 710072, China.}%
\thanks{$^{2}$University of Georgia, Athens, GA 30602, USA.}%
\thanks{*Corresponding author: Tianming Liu, tliu@cs.uga.edu; Shu Zhang, shu.zhang@nwpu.edu.cn.}
}

\begin{document}

\maketitle
\thispagestyle{empty}
\pagestyle{empty}

\begin{abstract}

Human-in-the-loop reinforcement learning integrates human expertise to accelerate agent learning and provide critical guidance and feedback in complex fields. However, many existing approaches focus on single-agent tasks and require continuous human involvement during the training process, significantly increasing the human workload and limiting scalability. In this paper, we propose HARP (Human-Assisted Regrouping with Permutation Invariant Critic), a multi-agent reinforcement learning framework designed for group-oriented tasks. HARP integrates automatic agent regrouping with strategic human assistance during deployment, enabling and allowing non-experts to offer effective guidance with minimal intervention. During training, agents dynamically adjust their groupings to optimize collaborative task completion. When deployed, they actively seek human assistance and utilize the Permutation Invariant Group Critic to evaluate and refine human-proposed groupings, allowing non-expert users to contribute valuable suggestions. In multiple collaboration scenarios, our approach is able to leverage limited guidance from non-experts and enhance performance. The project can be found at https://github.com/huawen-hu/HARP.

\end{abstract}

\section{INTRODUCTION}

In the field of multi-agent systems, reinforcement learning has shown great promise in fostering cooperation among agents, enabling them to solve complex tasks beyond the capabilities of individual agents \cite{kouzeghar2023multi, wang2024multi, feng2023variable, agrawal2023rtaw, igbinedion2024learning}. Research has shown that group division is an effective means of promoting collaboration, both in natural ecosystems \cite{wittemyer2007hierarchical} and in multi-agent systems \cite{phan2021vast} within artificial intelligence. Breaking teams into smaller units can facilitate more detailed learning processes while providing increased opportunities for integrating information-rich signals derived from group learning.  While reinforcement learning has proven effective for autonomous problem-solving and fostering cooperation among agents, it often struggles with low sample efficiency and poor generalization in intricate environments \cite{yu2018towards, zhang2024settling}.

Human-in-the-loop reinforcement learning (HITL-RL) represents a crucial advancement in overcoming the limitations of traditional reinforcement learning, particularly in complex multi-agent systems \cite{mandel2017add}. HITL-RL addresses these issues by incorporating human expertise directly into the learning process. Human intuition and domain knowledge provide essential guidance, enabling more accurate corrections in agent behavior and more effective integration of information-rich signals compared to fully automated methods. This collaboration between human insight and algorithmic power not only accelerates learning but also enhances the system's ability to generalize in complex tasks. As a result, HITL-RL offers a transformative approach where human input is not just a supplement, but a critical component for achieving higher performance and efficiency in real-world applications \cite{torne2023breadcrumbs, holk2024polite, mitra2024enhanced}.


\begin{figure}       
	\centering
        \includegraphics[scale=0.4]{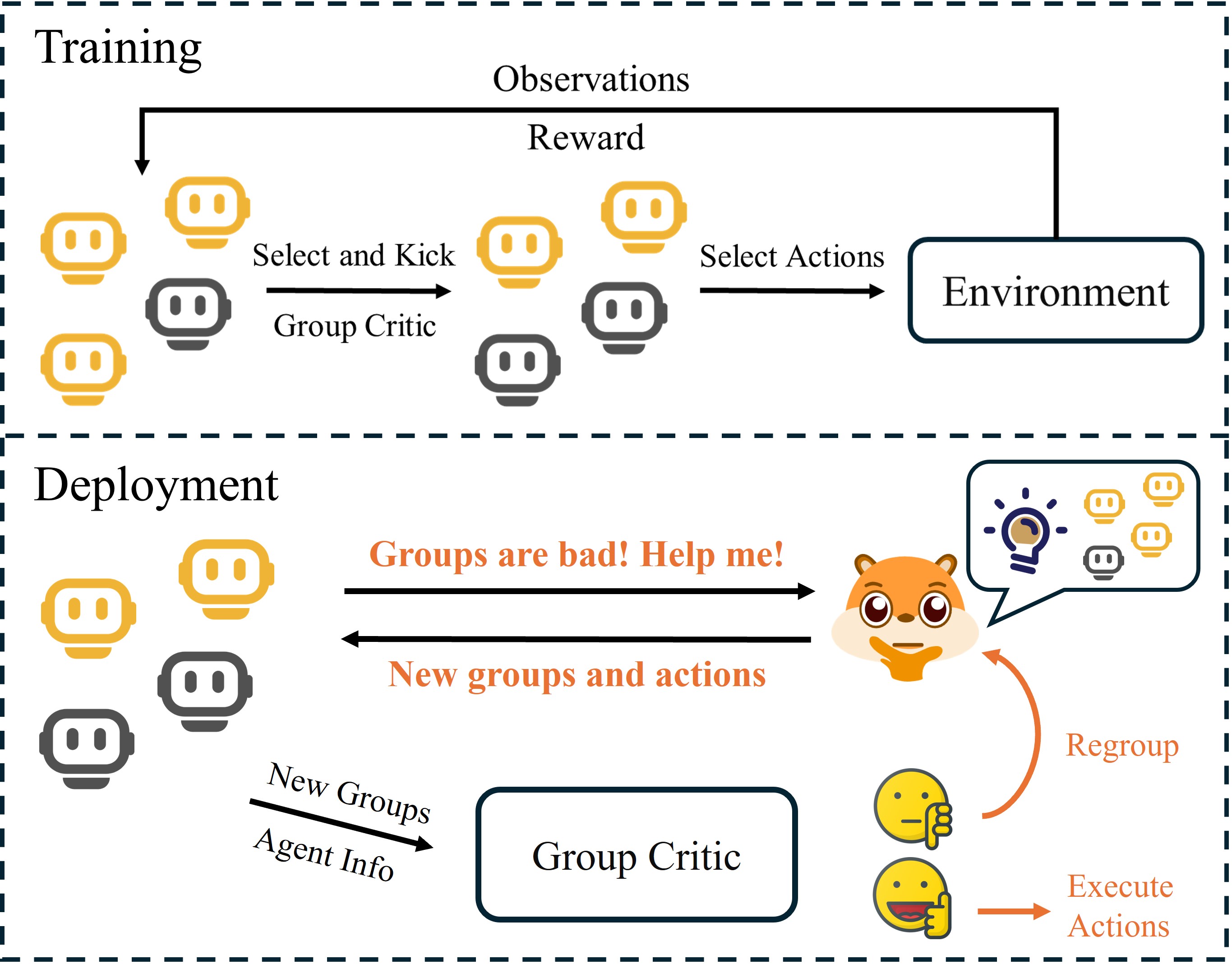}  
	\caption{HARP automatically forms groups during training to achieve collaborative task completion. In the deployment phase, it actively seeks assistance from humans, evaluates their suggestions, and provides feedback on the groups received.}   
	\label{fig: Grouping}
\end{figure}

Despite these advancements, existing human-in-the-loop methods predominantly focus on single-agent scenarios, while the integration of human guidance in multi-agent reinforcement learning settings remains largely unexplored. Extending these methods to multi-agent systems presents unique challenges, as guidance from human experts is expensive and rare. The core challenge lies in how humans can effectively provide guidance to multiple agents simultaneously, considering the complex dynamics and interactions within these systems \cite{chung2020battlesnake, retzlaff2024human, qin2024proactive}.

To address these challenges, we propose a novel framework for human-in-the-loop multi-agent reinforcement learning with dynamic grouping as shown in Fig. \ref{fig: Grouping}. Our approach introduces a mechanism for readjusting and reevaluating agent groupings during the deployment phase, guided by non-expert human input. It mitigates the burden of continuous human involvement in the reinforcement learning training process, enables non-expert humans to improve their suggestion-making skills over time through the reevaluation of their guidance, and enhances the adaptability of multi-agent systems to complex, dynamic environments. Our main contributions are as follows:

1. We propose a novel human-in-the-loop grouped multi-agent reinforcement learning framework where agents actively seek human assistance during the deployment phase, while requiring no human guidance during training.

2. We implement a permutation invariant grouping evaluation method that, during the deployment phase, utilizes non-expert human guidance to improve agent grouping and decision-making through regrouping and reevaluation.

3. Our experimental results on cooperative problems across three difficulty levels of StarCraft II \cite{samvelyan19smac} demonstrate that with limited human guidance during the deployment phase, HARP can significantly improve agent performance by over 10\%.

\section{RELATED WORK}

The advancements in multi-agent reinforcement learning (MARL) have significantly enhanced learning efficiency and performance through innovative approaches to agent grouping and role assignment. For instance, ROMA \cite{wang2020roma} introduces a role-oriented methodology that optimizes conditional mutual information to ensure precise alignment between roles and trajectories. RODE \cite{wang2020rode} expands upon this by deconstructing the joint action space and incorporating action effects into role policies. Unlike role-based methods, VAST \cite{phan2021vast} examines the influence of subgroups on value decomposition, employing variable subteams to aggregate local Q-functions into a group Q-function, which is then synthesized into a global Q-function via value function factorization. SOG \cite{shao2022self} proposes a dynamic grouping framework where designated commanders extend team invitations, allowing agents to select preferred commanders. GoMARL \cite{zang2024automatic} employs a "select and kick-out" strategy for automated grouping and integrates hierarchical control within policy learning. This approach achieves efficient cooperation without necessitating domain knowledge. Other grouping methods include approaches like GACG \cite{duan2024group}, which presents a graph-based technique that models the multi-agent environment as a graph, calculating cooperation demands between agent pairs and capturing dependencies at the group level.



Recently, human-in-the-loop reinforcement learning has emerged as a transformative approach to enhance policy learning efficiency. Various methodologies have been proposed to integrate human expertise into the learning framework. TAMER \cite{knox2008tamer} incorporates human experts into the agent's learning loop, allowing them to provide reward signals and minimizing discrepancies between the agent's policy and the human reinforcement function. COACH \cite{celemin2019interactive} enables non-expert humans to offer actionable advice during interactions with continuous action environments, using binary correction signals to refine agent actions. Recognizing the limited availability of human expert advice, RCMP \cite{da2020uncertainty} introduces a selective guidance strategy based on cognitive uncertainty. This approach requests human input only when the agent's uncertainty is high, employing specialized techniques to assess and quantify this uncertainty. The idea is further expanded by HULA \cite{singi2024decision}, which integrates human expert assistance during the deployment phase. In this approach, the agent actively seeks human recommendations when the return variance exceeds a predefined threshold. 
Recently, OpenAI's o1 model \cite{openai_o1} represents a significant advancement in natural language artificial intelligence (AI), leveraging human-in-the-loop reinforcement learning. By incorporating human feedback into its learning process, o1 demonstrates remarkable improvements in response accuracy, and adaptability across various tasks, showcasing the power of human-guided AI optimization.

In this paper, we propose a novel approach that integrates multi-agent reinforcement learning grouping techniques with human-in-the-loop learning. Our method introduces a mechanism for readjusting and reevaluating agent groupings during the deployment phase, leveraging guidance from non-expert humans. This approach mitigates the burden of continuous human involvement throughout the reinforcement learning training process. Furthermore, it enables non-expert humans to learn how to provide increasingly effective suggestions over time.

\section{METHOD}

\subsection{Preliminary}


In this paper, cooperative tasks are considered which involving $n$ agents, denoted as $A = \{a_1, ..., a_n\}$, framed as a decentralized partially observable Markov decision process (Dec-POMDP) \cite{oliehoek2016concise}. The process is defined by the tuple $G = \langle S, U, P, r, Z, O, n, \gamma \rangle$. The environment is characterized by a global state $s \in S$. At each time step $t$, each agent $a$ selects an action $u_a^t$ from its own action space $U_a$, forming a joint action $u^t \in U^n = U_1 \times \dots \times U_n$. The joint action determines the state transition according to the probability distribution $P(s^{t+1} | s^t, u^t): S \times U^n \times S \rightarrow [0, 1]$. A shared reward is provided by the function $r(s, u): S \times U^n \rightarrow R$, and future rewards are discounted by a factor $\gamma \in [0, 1)$.



\begin{figure*}       
	\centering
        \includegraphics[scale=0.6]{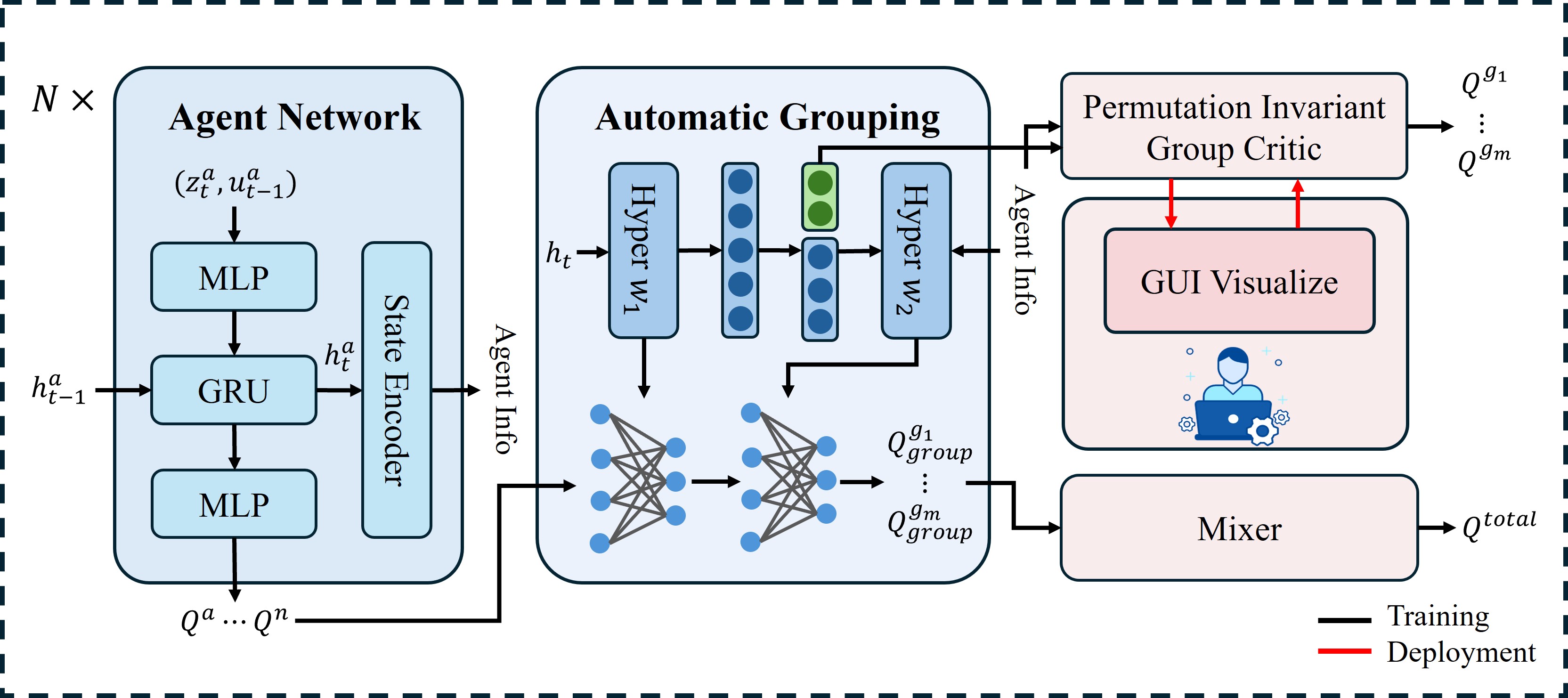}  
	\caption{The overall framework of HARP. The Agent Network uses gate recurrent unit (GRU) to capture long-term dependencies in past sequences and encodes hidden layer states to obtain state representations. The Automatic Grouping section utilizes Select and Kick along with hypernetworks to achieve dynamic grouping. The rightmost part shows the Mixer network and human participation component, including the Permutation Invariant Group Critic.}   
	\label{fig: Overview of HARP}
\end{figure*}


\subsection{Automatic Grouping Mechanism}

Consider a cooperative task involving a set of $n$ agents. We can partition these agents into a series of groups $G = \{g_1, \ldots, g_m\}$, where $1 \leq m \leq n$. Each group $g_j$ contains a subset of agents: $g_j = \{a_{j}^{1}, \ldots, a_{j}^{n_j}\} \subseteq A$. The union of all groups covers the entire set of agents: $\bigcup_{j=1}^m g_j = A$. Also, the groups are mutually exclusive, meaning $g_j \cap g_k = \emptyset$ for $j, k \in \{1, 2, \ldots, m\}$ and $j \neq k$.

In this section, we focus on developing an effective learning method for dynamic group adjustment. Our objective is to learn a grouping function $f_g: A \rightarrow G$ that maps agents to groups.

To achieve this, we introduce an automatic grouping mechanism, as illustrated in Fig. \ref{fig: Overview of HARP}. Following the value function decomposition approach \cite{sunehag2018value, son2019qtran}, we represent the group Q-value $Q_g$ as an aggregation of individual agent Q-values $Q_g^i$ within the group:


\begin{equation}
Q_g = \text{Aggregate}(Q_g^1, Q_g^2, \ldots, Q_g^{n_j})
\end{equation}

This mechanism allows agents to dynamically adjust their groupings based on the learned Q-values, potentially leading to more effective cooperation in complex tasks.

Given the hidden state $h_t^a$ for each agent, we employ a group weight generator (hyper $w_1$) to learn the contribution $w_1$ of each agent to the total Q-value of the current group. We then apply a ``Select and Kick'' strategy to adjust the groupings. Consider two groups $g_1$ and $g_2$, with a set of weights $w_1 = \{\{w_1^i\}, \{w_1^j\}\}$, where $i$ represents agents in group $g_1$ and $j$ represents agents in group $g_2$. We first calculate the threshold $\tau_1$ for group $g_1$ as the average of $\{w_1^i\}$:
\begin{equation}
\tau_1 = \frac{1}{|g_1|} \sum_{i \in g_1} w_1^i
\end{equation}

This threshold is then used to reassign agents between groups. Agents in $g_1$ with weights below $\tau_1$ are moved to $g_2$, resulting in updated group compositions:
\begin{align}
g_1' &= \{i \in g_1 \mid w_1^i \geq \tau_1\} \\
g_2' &= g_2 \cup \{i \in g_1 \mid w_1^i < \tau_1\}
\end{align}

The process is then repeated for the updated group $g_2'$. This dynamic process adjusts group compositions based on each agent's contribution to the group, potentially improving overall system performance in multi-agent scenarios.

After obtaining group indices based on $w_1$, we utilize hyper $w_2$ to generate intra-group feature weights. For each group, we derive the group state by pooling the agent information within the group and apply a multi-layer perceptron to compute the Q-value for each group. Subsequently, we employ a Mixer \cite{zang2024automatic} to transform these group-level Q-values into an overall Q-value.

\subsection{Permutation Invariant Group Critic}

In multi-agent reinforcement learning scenarios, each agent $i$ in a system possesses a unique state representation $\mathbf{s}_i$. A critical challenge is to compute an accurate joint value function $V(\mathbf{s}_1, \mathbf{s}_2, \ldots, \mathbf{s}_n)$ based on these individual states. Conventionally, this is achieved by concatenating the states into a single vector:
\begin{equation}
\mathbf{s} = [\mathbf{s}_1, \mathbf{s}_2, \ldots, \mathbf{s}_n]
\end{equation}

This vector is then input into the critic function. However, this method is sensitive to the ordering of agent states, leading to the permutation non-invariance problem:
\begin{equation}
\text{Critic}([\mathbf{s}_1, \mathbf{s}_2, \ldots, \mathbf{s}_n]) \neq \text{Critic}(\pi([\mathbf{s}_1, \mathbf{s}_2, \ldots, \mathbf{s}_n]))
\end{equation}
where $\pi(\cdot)$ denotes any permutation of the state sequence. This issue is even more critical in grouping cause both within-group permutation invariance and between-group permutation invariance need to be satisfied.

\begin{figure}       
	\centering
        \includegraphics[scale=0.48]{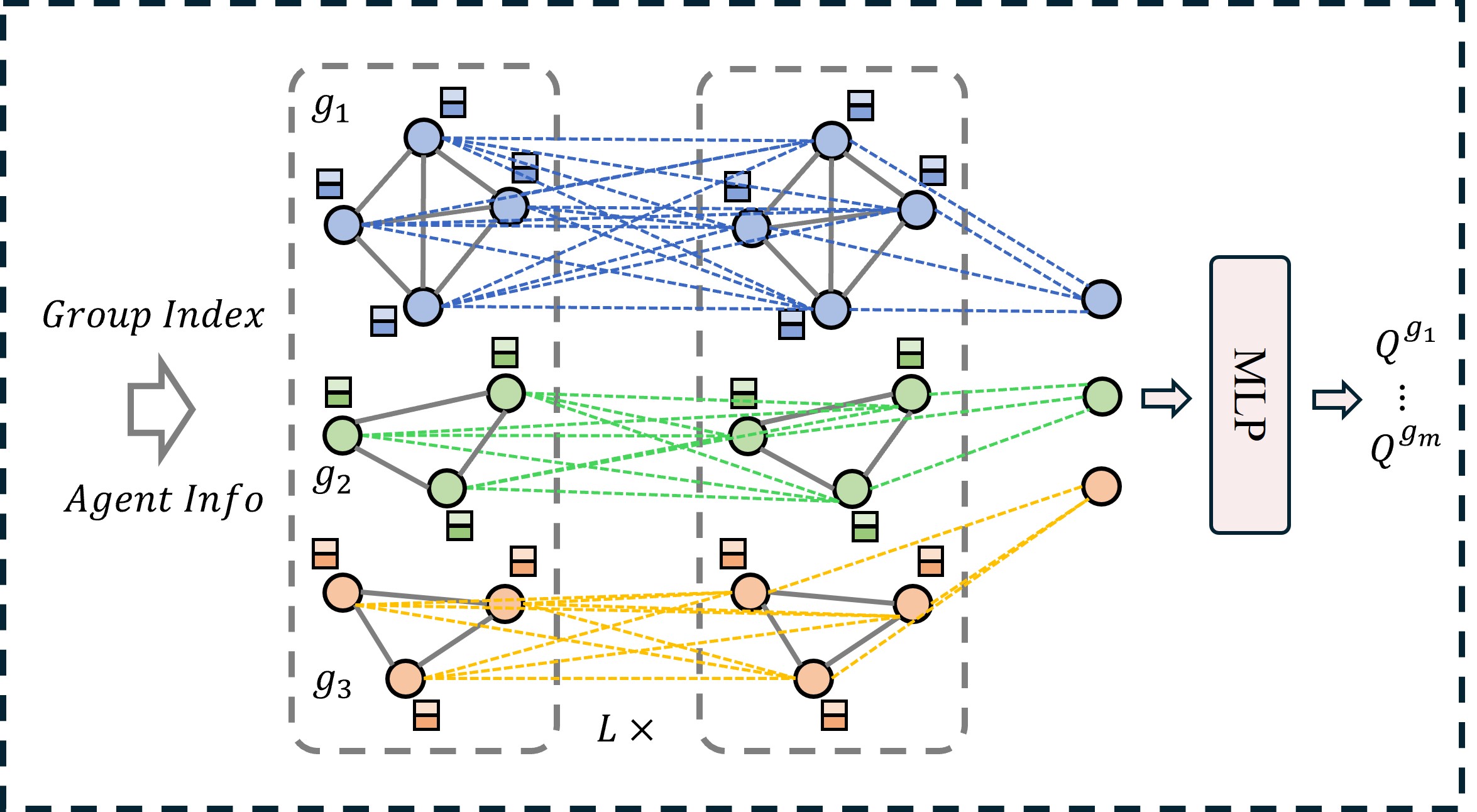}  
	\caption{Permutation Invariant Group Critic}   
	\label{fig: Permutation Invariant Group Critic}
\end{figure}

Inspired by Liu et al. \cite{liu2020pic}, we propose the Permutation Invariant Group Critic (PIGC). In this approach, the multi-agent environment is represented as a graph, where agents are modeled as nodes and their interactions as edges. Given the group index information, we construct the graph adjacency matrix, where for agents within the same group, we construct an edge between any two agents, and for agents in different groups, we do not construct any connections, i.e. each group is an independent subgraph. Meanwhile, we encode the hidden states of each agent as the node embedding of the graph, as shown in Fig. \ref{fig: Permutation Invariant Group Critic}.

To compute the output of the group critic, we use a L-layer graph convolutional network $GCN = \{f_{GCN}^{(1)}, \dots, f_{GCN}^{(L)}\}$. Graph convolutional network (GCN) layer processes input data in the form of node features and the graph's connectivity structure, typically represented by an adjacency matrix. 



\begin{equation}
h^{(l)}=f_{GCN}^{(l)}(h^{(l-1)}):=\sigma(\hat{A}_{adj}h^{(l-1)}W^{(l)})
\end{equation}
where $\hat{A}_{adj}=A_{adj}+\boldsymbol{I}_N$ is the graph adjacency matrix with self-connections, $\boldsymbol{I}_N$ is the identity matrix, $W^{(l)}$ is a weight matrix, $\sigma$ is the activation function. Then we use a fully connected layer to obtain the Q value of each group. These Q-values are compared with the group-level Q-values obtained from the automatic grouping process, and their L2 loss is computed. This loss is incorporated as a component of the overall loss function.

\subsection{Human Participation in Group Adjustment}

Relying solely on automatic grouping during training has limited performance at deployment time. In dynamic scenarios, while fixed groupings may achieve good performance during training, a single grouping strategy cannot satisfy all states. Let's use playing soccer as an example. Although we can train players to automatically learn teamwork and complete tasks, seeking optimal groupings, in actual matches (i.e., the deployment phase), a single grouping strategy cannot work effectively from start to finish. Therefore, a coach needs to pause the game appropriately and make timely adjustments to the grouping strategy.


\begin{algorithm}
    \caption{HARP Deployment}
    \begin{algorithmic}
        \WHILE {not terminate}
            \STATE Get $h_t$ and $Q^a$ using agent network

            \STATE Load group index $G$
            
            \STATE Compute variance of group $Var(G)$ using Eq. (8)
            
            \IF{$Var(G)$ $\ge$ $\epsilon$}

                \WHILE {not done}
            
                    \STATE Get new group index $G_{human}$ and actions $u_{human}$ from human

                    \STATE Get new group Q values $Q^{G_{human}}$ using Permutation Invariant Group Critic

                    \IF{$Q^{G_{human}} >  Q^G$} 
                        
                        \STATE done

                    \ENDIF

                \ENDWHILE
                \STATE Execute $u_{human}$

            \ELSE

                \STATE Execute actions $u=argmax(Q^a)$
            
            \ENDIF
        
        \ENDWHILE
    \end{algorithmic}
\end{algorithm}

In this paper, we adopt a proactive approach to seeking help. During the deployment phase, we define "the variance of group return". When this variance exceeds the maximum value in the historical queue, the agents actively seek human assistance to adjust the grouping in a timely manner. However, guidance from human experts is expensive and rare. To address this, we propose a regrouping and reevaluation strategy as shown in the red line in Fig. \ref{fig: Overview of HARP} that fully utilizes non-expert human guidance. By reevaluating non-expert human guidance, we can reasonably identify shortcomings in the current grouping strategy, provide timely feedback to humans, and make adjustments. This achieves a two-way learning process where the agent gets help while humans can also learn how to come up with more effective groupings.

\begin{equation}
Var = \alpha\sum_{g \in G} \frac{1}{|g|} \sum_{a \in g} (Q_g^a - \overline{Q_g})^2 + \beta\frac{1}{|G|} \sum_{g \in G} (Q_g - \overline{Q})^2     
\end{equation}
where $\alpha$ is the coefficient for intra-group variance, and $\beta$ is the coefficient for inter-group variance.

We comprehensively consider both intra-group and inter-group variances as the basis for determining whether assistance is needed. By setting coefficients for intra-group and inter-group variances, we can effectively adjust the composition of agent teams. In homogeneous agent tasks, we focus more on both intra-group and inter-group variances. In heterogeneous agent tasks, we place greater emphasis on inter-group variance. This approach allows us to tailor our grouping strategy to the specific characteristics of the task at hand.  When dealing with both intra-group and inter-group variances simultaneously, we normalize each component using values from the historical variance queue.

\section{RESULTS}
\subsection{Experiment Settings}

We conducted our experiments on six maps in the StarCraft II Multi-Agent Challenge environment, encompassing three difficulty levels: Easy (8m, MMM), Hard (8m\_vs\_9m, 5m\_vs\_6m), and Super Hard (MMM2, corridor). Agents controlled by algorithms compete against those controlled by systems to gain rewards. The one that eliminates the opponent first achieves victory. Throughout the training process, we maintained consistent parameter settings and the length of historical queue was set to 10. Our experiments were executed on an NVIDIA GeForce RTX 3090 GPU with 24GB memory. 

\begin{figure*}       
	\centering
        \includegraphics[scale=0.29]{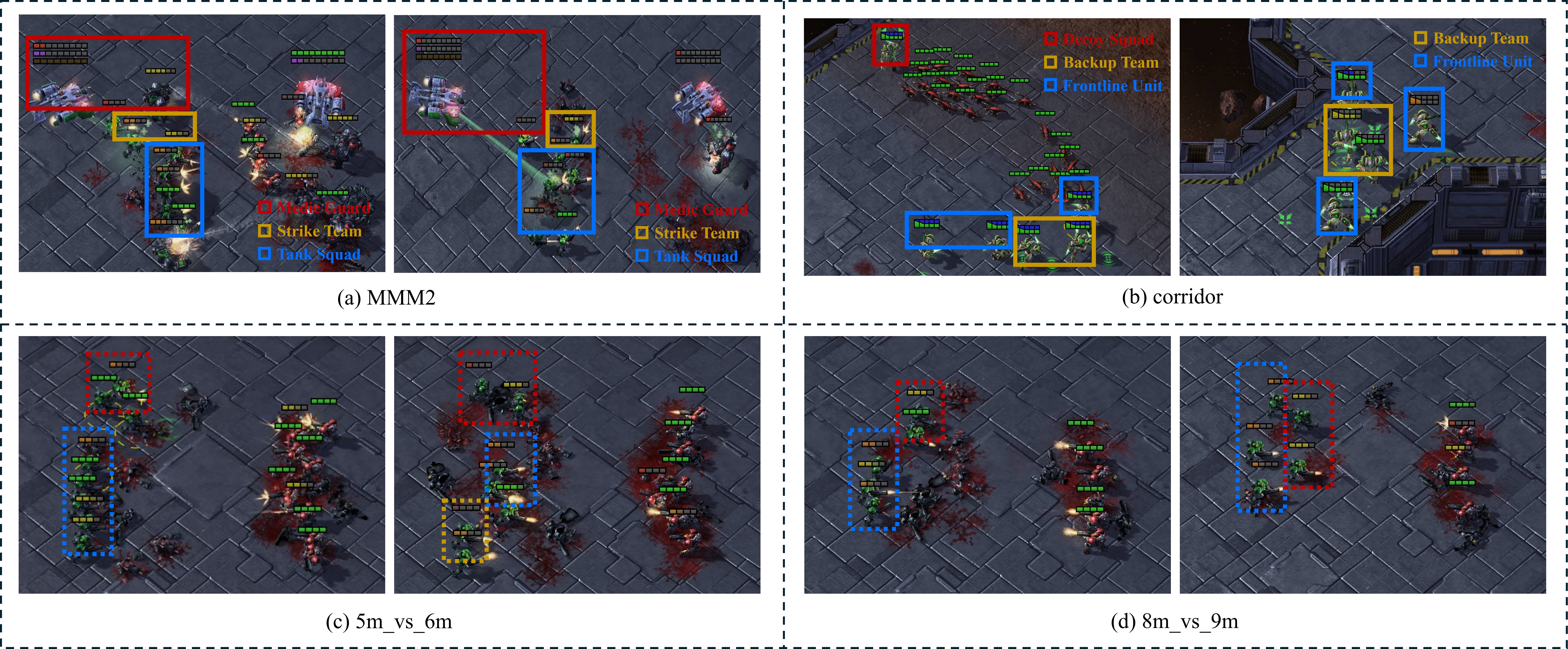}  
	\caption{Grouping visualization during training and deployment phases. (a) and (b) show the visualization and interpretability analysis of automatic grouping results during the training process, while (c) and (d) present the visualization of human-assisted results during the deployment phase. The 'm' in these maps refers to Marine. 'MMM' represents a battle configuration of 1 Medivac, 2 Marauders, and 7 Marines on each side, while 'MMM2' with 1 Medivac, 2 Marauders, and 7 Marines against 1 Medivac, 3 Marauders, and 8 Marines. In the 'corridor' map, players control 6 Zealots facing 24 Zerglings.}   
	\label{fig: maps-analysis}
\end{figure*}

\subsection{Comparison with Other Methods}

We selected VAST and GACG as our comparison methods. VAST is similar to our work, which is also based on value decomposition and transforms local Q-functions of individual subgroups into group Q-functions through linear aggregation. GACG employs a graph theory algorithm to achieve agent grouping. We also compared a baseline model, HARP without human feedback (Baseline), which uses a greedy strategy during training by selecting the action with the highest Q-value at each step. 

\begin{figure}       
	\centering
        \includegraphics[scale=0.55]{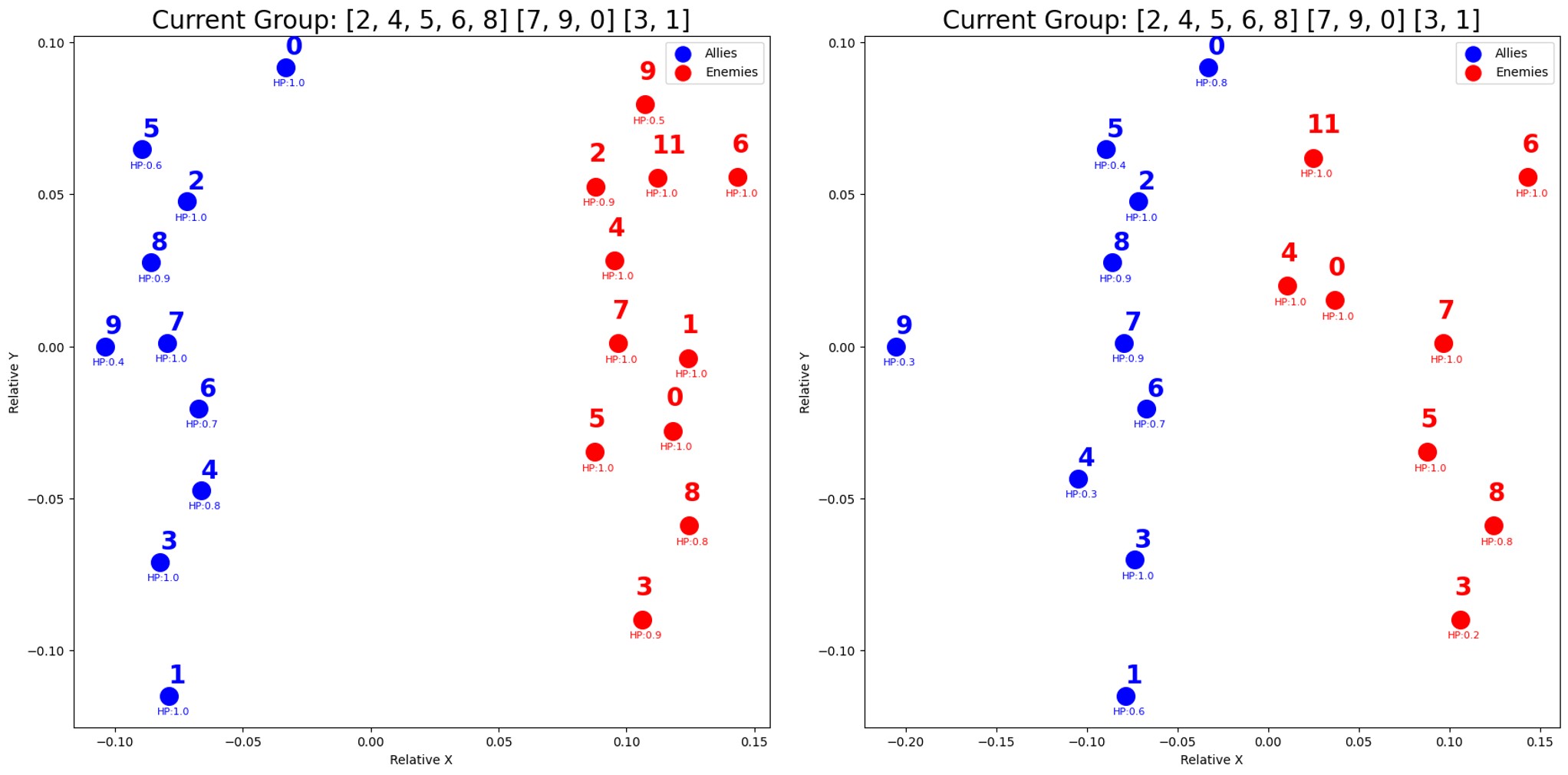}  
	\caption{The GUI interface shown to humans during the deployment phase, including the relative positions of each agent and their health percentages.}   
	\label{fig: GUI of MMM2}
\end{figure}

We compared the test win rate and the average return across these different methods. As shown in Fig. \ref{fig: GUI of MMM2}, to enable humans to better guide the behavior of the multi-agent system, we implemented a graphical user interface (GUI) that displays partial information about the environment, including the relative positions between agents and their health status. Based on this information, humans can determine how to improve the grouping strategy and provide guidance for the agents' actions. The experimental results are shown in Table \ref{table: test win rate} and Table \ref{table: mean return}.

\begin{table}[H]
\caption{Comparison of HARP with other methods on test win rate (\%).}
\setlength{\tabcolsep}{1.5mm}{
\centering
\begin{tabular}{lccccccc}
\hline
Method & 8m & MMM & 5m\_vs\_6m & 8m\_vs\_9m & MMM2 & Corridor \\
\hline
VAST & 98.1 & 94.4 & 71.2 & 91.9 & 43.7 & 0 \\
GACG & 98.7 & 98.7 & 53.1 & 90.6 & 48.7 & 0 \\
Baseline & 96.0 & \textbf{100} & 65.6 & 90.6 & 93.7 & 93.7 \\
HARP & \textbf{100} & \textbf{100} & \textbf{100} & \textbf{100} & \textbf{100} & \textbf{100} \\
\hline
\end{tabular}
}
\label{table: test win rate}
\end{table}

\begin{table}[H]
\caption{Comparison of HARP with other methods on test mean return.}
\setlength{\tabcolsep}{1.5mm}{
\centering
\begin{tabular}{lccccccc}
\hline
Method & 8m & MMM & 5m\_vs\_6m & 8m\_vs\_9m & MMM2 & Corridor \\
\hline
VAST & 19.8 & 19.7 & 17.2 & 19.4 & 15.8 & 11.0 \\
GACG & \textbf{20.7} & \textbf{21.7} & 15.5 & 19.3 & 16.3 & 10.9 \\
Baseline & 19.8 & 20.3 & 16.5 & 19.3 & 19.3 & 19.8 \\
HARP & 20.0 & 20.1 & \textbf{20.0} & \textbf{20.0} & \textbf{20.1} & \textbf{20.2} \\
\hline
\end{tabular}
}
\label{table: mean return}
\end{table}

On easy-type tasks, each method achieves satisfactory performance, with win rates approaching or reaching 100\%. However, on hard-type tasks, the performance gap widens significantly. On the 5m\_vs\_6m map, methods other than HARP only achieve win rates of around 50\% to 70\%. Interestingly, on the 8m\_vs\_9m map, this gap narrows. Although both maps present a scenario where allies are outnumbered by one, the performance on 8m\_vs\_9m is notably better than on 5m\_vs\_6m. We posit that as the number of agents increases, the model becomes more adept at grouping, allowing for clearer task allocation within each group. On super hard-type maps, only the baselines and HARP manage to maintain high success rates, while VAST and GACG perform poorly or even fail to function. Notably, HARP achieves a 100\% win rate across all six maps of varying difficulties.

Fig. \ref{fig: maps-analysis} (a) and Fig. \ref{fig: maps-analysis} (b) show partial results from the training process, which we analyzed for interpretability based on the learned groupings. It can be observed that different groupings exhibit distinct strategies. On the MMM2 map, the grouping consists of three parts: based on their performance in the game, we can categorize them as the Medic Guard which protect the Medivac, the Tank Squad responsible for drawing fire and attacking, and a small group consisting of only two Marines (Strike Team). This Strike Team switches between the roles of the other two; when the number of Medic Guard decreases, the Strike Team takes on part of the responsibility for protecting the Marines. A similar result can be observed on the corridor map, where the Decoy Squad is responsible for drawing the majority of the Zerglings, while the Backup Team and Frontline Unit handle the small portion of enemies not drawn away. Among these, the Frontline Unit is primarily responsible for attacking, and the Backup Team finishes off any remaining enemies.

\subsection{Human Participation Rate in Deployment Phase}

\begin{figure}       
	\centering
        \includegraphics[scale=0.4]{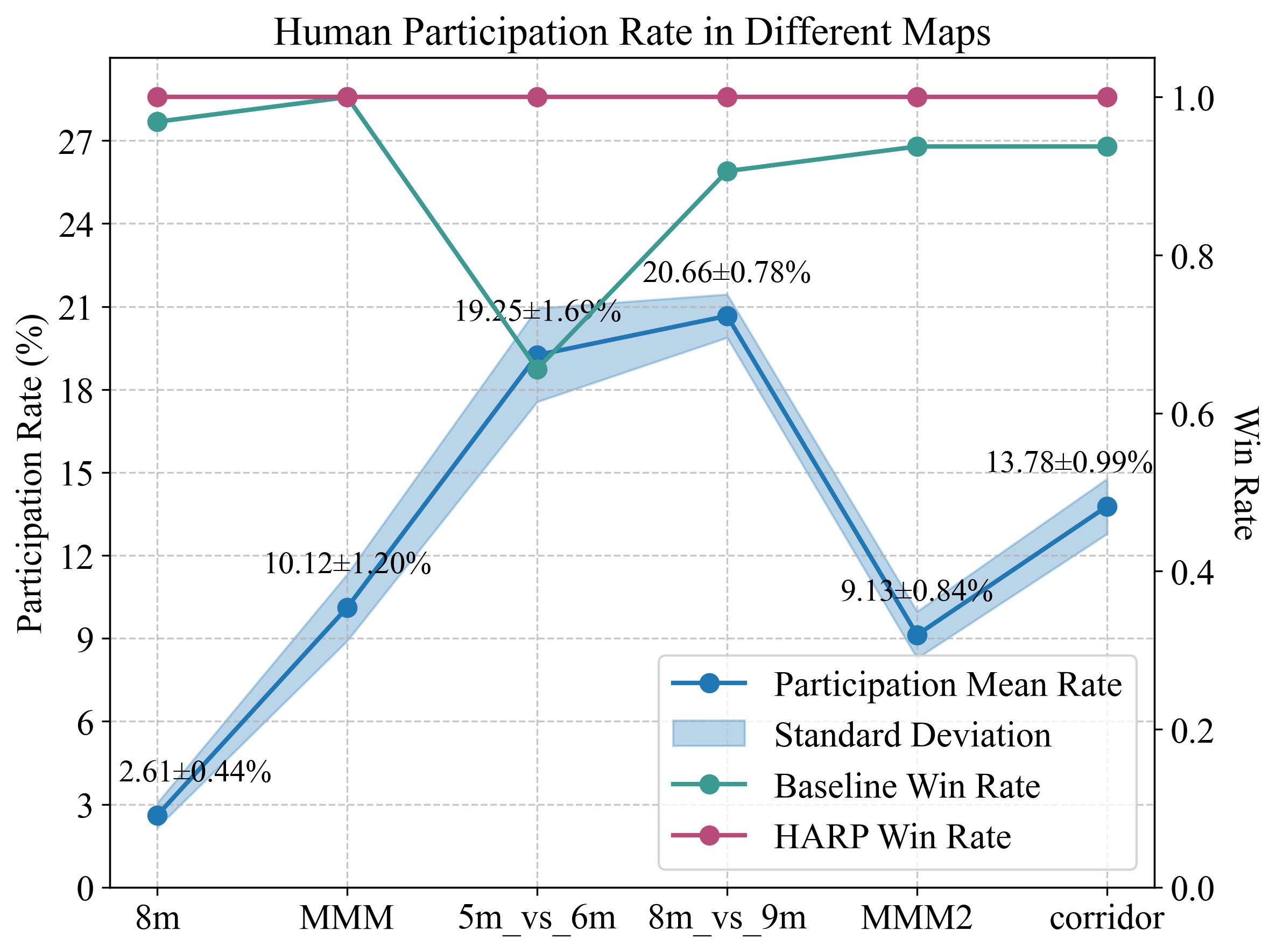}  
	\caption{The number of times agents actively seek human assistance during the deployment phase on different types of maps, expressed as a percentage (total human interventions / total steps).}   
	\label{fig: Human's Participation}
\end{figure}

In this section, we investigated the proportion of human involvement during the deployment phase across different types of maps. We repeated the experiments five times on each map, with each experiment consisting of 31 episodes. We recorded the total number of steps and the number of human interventions, using their ratio as an indicator of human involvement. The results are shown in Fig. \ref{fig: Human's Participation}. Overall, the proportion of human-assisted agent reorganization across all maps was less than 25\%. On simpler maps, the agents requested human assistance less frequently, accounting for only 2.61\% on the 8m map. However, in tasks that agents did not handle well, such as 5m\_vs\_6m and 8m\_vs\_9m, where the agents' win rates were only 65.6\% and 90.6\%, respectively, the number of times agents sought human assistance significantly increased during the deployment phase, rising to around 20\%. With HARP's limited human involvement in the deployment phase, we can significantly improve the agent's performance.

\subsection{Impact of Human Assistance on Challenging Scenarios}

\begin{table}[H]
\caption{Group information learned through automatic grouping}
\setlength{\tabcolsep}{11.5mm}{
\centering 
\begin{tabular}{lc}
\hline
 Maps & Groups  \\
\hline
8m &  [0 1 3 4 5] [6 7] [2] \\
MMM & [0 1 2 3 4 5] \\
5m\_vs\_6m & [0 1 2 3 4] \\
8m\_vs\_9m & [0 1 2 3 4 5 6 7] \\
MMM2 & [2 4 5 6 8] [7 9 0] [3 1]\\
corridor & [1 2] [0 4 5] [3] \\
\hline
\end{tabular}
}
\label{table: groups}
\end{table}

In this section, we examine why limited human guidance during the deployment phase significantly improves the success rate of the game. As observed in Table \ref{table: test win rate}, all baselines perform poorly on the 5m\_vs\_6m map, showing a substantial performance gap compared to the 8m\_vs\_9m map of similar difficulty. However, actively seeking human assistance during the testing phase leads to a marked improvement in performance. We aim to understand the specific role of human assistance in this context.

Table \ref{table: groups} outlines the groupings learned by the automatic grouping algorithm across various maps. On the 5m\_vs\_6m and 8m\_vs\_9m maps, the automatic grouping algorithm places all agents in a single group for coordination and cooperation. This approach results in a lack of clear division of labor among agents. As illustrated in Fig. \ref{fig: maps-analysis} (c) and Fig. \ref{fig: maps-analysis} (d), the dashed lines represent human-assisted groupings. During testing, there is evident division of labor and strategy among agents. At certain moments, agents exhibit spatial convergence, forming groups. Simultaneously, in terms of strategy, agents with higher health points position themselves more forward, while those with lower health points retreat. This phenomenon is also observable in Fig. \ref{fig: GUI of MMM2}. During the human assistance process, these phenomena and strategies are typically leveraged to dynamically adjust the behavior and grouping of multiple agents. This adaptive approach, guided by human insight, appears to be a key factor in the improved performance observed when human assistance is incorporated into the deployment phase.



\section{CONCLUSIONS}

In this work, we propose an effective multi-agent reinforcement learning framework that actively seeks human assistance for grouping during the deployment phase. As real-time human involvement during training can be cumbersome and time-consuming, we shift human participation to the deployment stage. To enhance the quality of non-expert guidance, we introduce a regrouping and reevaluation method for group critics based on group invariance. By evaluating human-proposed groupings, we maximize the utilization of human suggestions. We tested our approach on six StarCraft II maps across three difficulty levels. Compared to scenarios without human assistance, our method improved success rates by an average of 10\%. On some more challenging maps, we increased success rates from 65\% to 100\%. Our approach has the potential to be extended to other tasks such as human-machine collaboration and sim-to-real transfer \cite{liu2023digital, liu2023dt, wu2023human, abeyruwan2023sim2real}. One promising direction is its integration with multimodal LLMs for complex reasoning tasks \cite{wang2024large}. By dynamically grouping agents to specialize in different data modalities and adaptively regrouping based on task demands, the framework enhances multimodal alignment and balances the permutations of inputs. Incorporating real-time human feedback further improves decision-making, benefiting applications such as autonomous vehicles processing diverse sensor data or healthcare systems integrating various patient information. This integration helps make multimodal LLMs more robust and adaptable to complex, real-world challenges.










\bibliographystyle{IEEEtran}
\bibliography{ref}

\end{document}